\documentclass[sigconf]{acmart}

\usepackage[tableposition=top]{caption}
\usepackage{subcaption}
\usepackage{float}

\AtBeginDocument{%
  \providecommand\BibTeX{{%
    \normalfont B\kern-0.5em{\scshape i\kern-0.25em b}\kern-0.8em\TeX}}}

\setcopyright{rightsretained} 

\acmSubmissionID{1154}

\begin{document}

\title{Multimodal Icon Annotation For Mobile Applications}

\author{Xiaoxue Zang}
\authornote{Both authors contributed equally to this research. Order is determined alphabetically.}
\email{xiaoxuez@google.com}

\affiliation{
  \institution{Google Research}
  \city{Mountain View}
  \state{CA}
  \country{United States}
}

\author{Ying Xu}
\authornotemark[1]
\email{yingxuv@google.com}
\affiliation{
  \institution{Google Research}
  \city{Mountain View}
  \state{CA}
  \country{United States}
}

\author{Jindong Chen}
\email{jdchen@google.com}
\affiliation{
  \institution{Google Research}
  \city{Mountain View}
  \state{CA}
  \country{United States}
}

\begin{abstract}
Annotating user interfaces (UIs) that involves localization and classification of meaningful UI elements on a screen is a critical step for many mobile applications such as screen readers and voice control of devices. Annotating object icons, such as menu, search, and arrow backward, is especially challenging due to the lack of explicit labels on screens, their similarity to pictures, and their diverse shapes. Existing studies either use view hierarchy or pixel based methods to tackle the task. Pixel based approaches are more popular as view hierarchy features on mobile platforms are often incomplete or inaccurate, however it leaves out instructional information in the view hierarchy such as resource-ids or content descriptions. We propose a novel deep learning based multi-modal approach that combines the benefits of both pixel and view hierarchy features as well as leverages the state-of-the-art object detection techniques. In order to demonstrate the utility provided, we create a high quality UI dataset by manually annotating the most commonly used 29 icons in Rico, a large scale mobile design dataset consisting of 72k UI screenshots. The experimental results indicate the effectiveness of our multi-modal approach. Our model not only outperforms a widely used object classification baseline but also pixel based object detection models. Our study sheds light on how to combine view hierarchy with pixel features for annotating UI elements.

\end{abstract}

\copyrightyear{2021}
\acmYear{2021}
\acmConference[MobileHCI '21]{23rd International Conference on Mobile Human-Computer Interaction}{September 27-October 1, 2021}{Toulouse \& Virtual, France}
\acmBooktitle{23rd International Conference on Mobile Human-Computer Interaction (MobileHCI '21), September 27-October 1, 2021, Toulouse \& Virtual, France}
\acmDOI{10.1145/3447526.3472064}
\acmISBN{978-1-4503-8328-8/21/09}



\begin{teaserfigure}
\centering
  \includegraphics[width=0.8\textwidth]{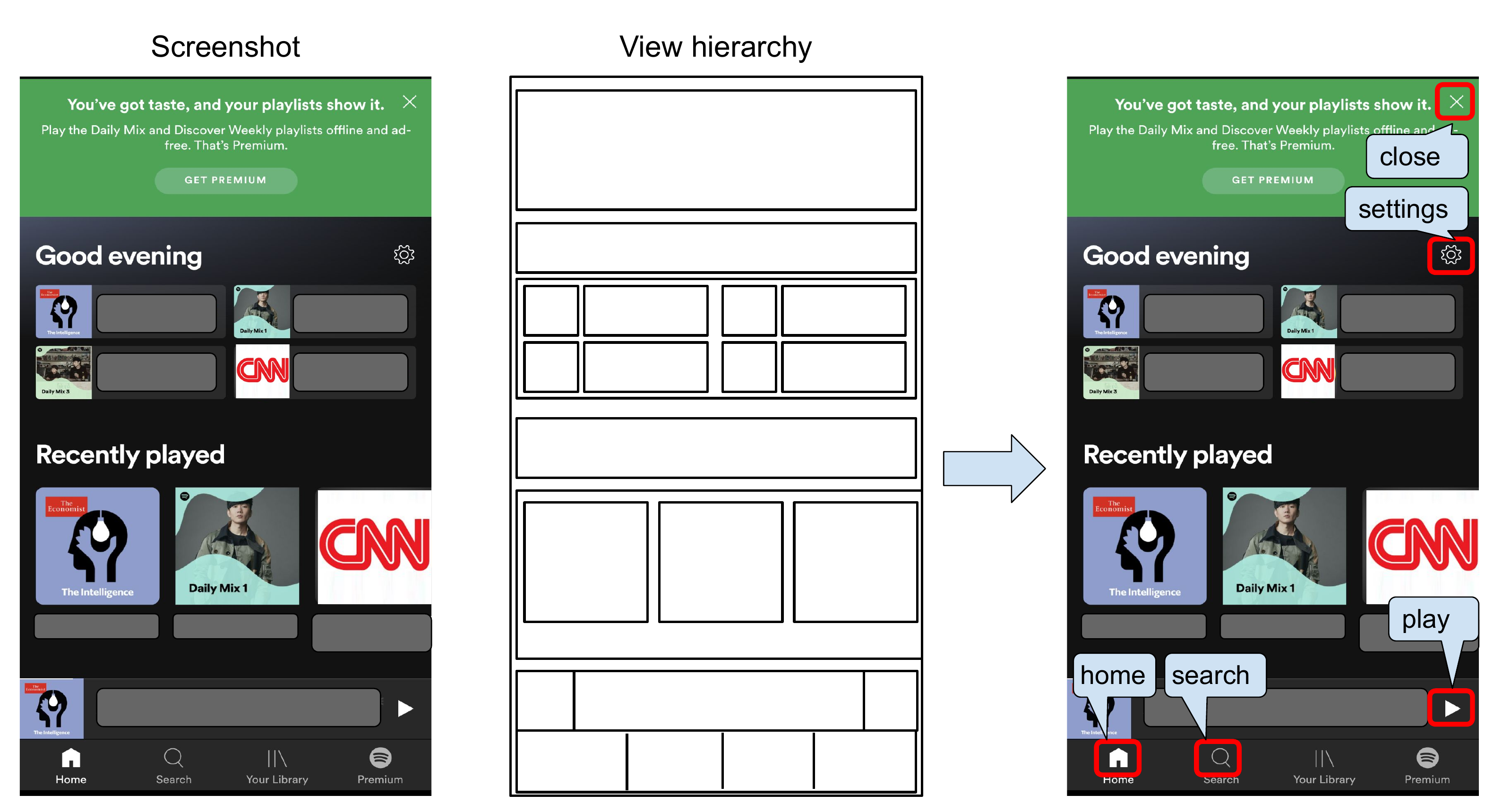}
  \caption{Our work proposes a multimodal deep-learning approach for annotating icons on a mobile UI. Given a screenshot and its view hierarchy, the model localizes and classifies 29 icons that are most commonly used in the daily life.}
  \Description{Three figures. First shows a screenshot of a music app. Second visualizes its corresponding view hierarchy. The third figure is the first screenshot annotated with icons.}
  \label{fig:teaser}
\end{teaserfigure}

\maketitle

\section{Introduction}


User interface (UI) element annotation is a task that localizes meaningful UI elements on a screen and classify their types. It is a critical task in mobile UI understanding and powers many accessibility applications such as TalkBack, a screen reader in Android, and Voice Access, which enables voice control of the phone \cite{li2020-mapping, szpiro2016people}.



\begin{figure}[t]
  \centering
  \includegraphics[width=\linewidth]{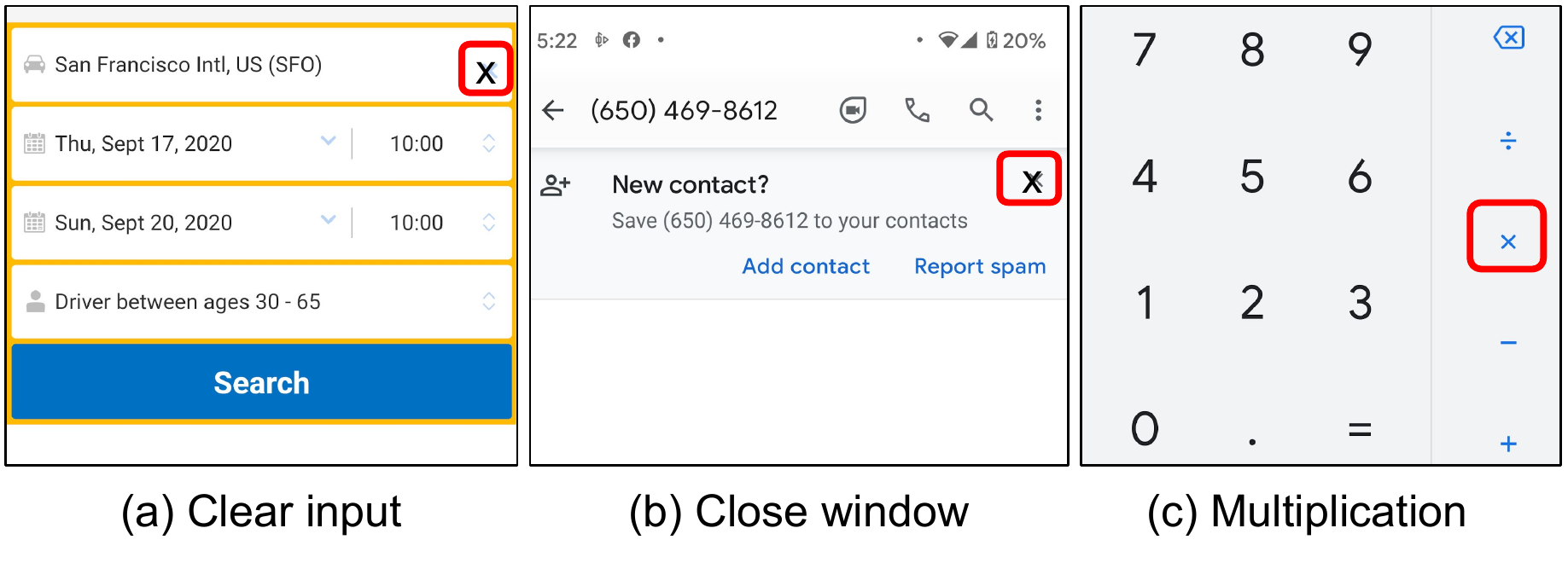}
  \caption{An example showing visually similar icons can have distinguishing functionalities.}
  \Description{Three cropped UIs that have x-shaped icons. Their functions are respectively clearing input, closing window, and doing multiplication.}
  \label{fig:similar_icons_example}
\end{figure}

The act of annotating object icons, that are graphical symbols representing a function such as arrow forward, more, and menu, is especially challenging due to the lack of explicit textual labels of icons on screens, their similarity to pictures and variations in their appearance. Due to this variation, icons of the same class don't necessarily have a unified appearance, and that similarly shaped icons aren't semantically similar. An example is shown in Figure~\ref{fig:similar_icons_example} where ``X" shaped icon means differently (e.g. clear inputs, close the prompt, or multiplication) under different scenarios. All of the aforementioned challenges make simple image patch matching method infeasible for the icon annotation task.

\begin{figure}[t]
  \centering
  \includegraphics[width=\linewidth]{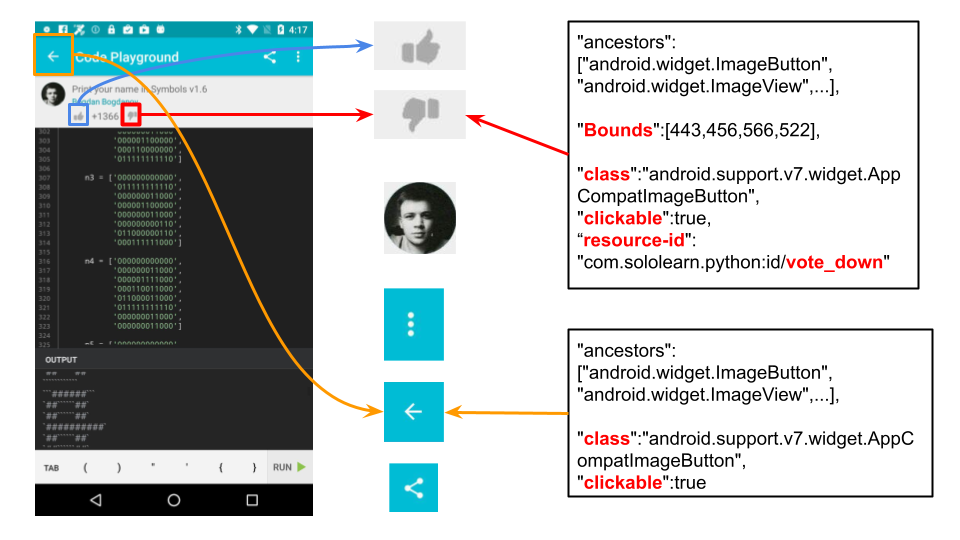}
  \caption{An example of app screenshot, UI elements and their view hierarchy information.}
  \Description{A screenshot that have some of its icon elements highlighted.  A part of view hierarchy information that describes the thumb down and arrow back icons on the screenshot are also shown.}
  \label{fig:screenshot_example}
\end{figure}

Existing studies for UI element annotation either use view hierarchy based or pixel based methods. Figure~\ref{fig:screenshot_example} shows a UI with both pixel and view hierarchy features. Though pixel features reveal the shape of icons which plays an important role in determining their semantic types, view hierarchy attributes, such as class name, resource-id, and content descriptions, also contain useful information for the task. As shown in Figure~\ref{fig:screenshot_example}, “resource-id” of thumbs down icon contains keywords ``vote\_down”, which serve as a useful signal implying its semantics. Thus, it is desirable to utilize both features to achieve better model performance.

Moreover, prior work usually tackles the icon annotation task in two steps. For instance, \citet{Liu:2018} first utilized the layout information in the view hierarchy to localize the icon elements and then adopted a pixel based object classification model to classify icon types. One problem of their approach is that view hierarchies may not precisely describe the UIs. The noise in view hierarchies leads to prediction errors of the model. Thus, \citet{zhang2021screen} proposed a pixel based pipeline. They first used a pixel based object detection model to identify UI elements of different categories, such as icons, texts, pictures, and checkboxes. Then for UI elements that are identified as icons, another classification model is used to predict their fine-grained types. However, they also suffer from the issue that the errors in the first stage propagate to the second stage.

In this paper, we propose a multi-modal (image and text modality) single stage object detection model for icon annotation. It fuses pixel features and text attributes in view hierarchies such as resource-ids, content descriptions, and class names. 
Our model has several advantages over previous approaches. First, as an object detection model, it takes the whole UI screenshot as input, thus can leverage more contextual information compared with classification models. Taking Figure \ref{fig:similar_icons_example}(c) as an example, if the ``X" shaped icon co-occurs with numerical numbers, it's more likely to mean multiplication than close or clear. However, such information is unavailable in the classification models as they only take the cropped UI element image as the input. Second, by fusing view hierarchy features with pixel features, the model is robust to noise in view hierarchies while taking advantage of the meaningful text features. 

The performance of our model and baselines are evaluated on a large-scale mobile app design dataset Rico~\cite{Deka:2017}, which contains more than 72k unique UI screens and their view hierarchies. 
However, the locations of icons in the original Rico data are proposed by bounding boxes in the view hierarchy and their types are labeled automatically by an image classification model, which inevitably brings errors and systematic bias~\cite{Liu:2018}. A recent study also shows that roughly only 10\% of the UIs can be deemed as high-quality design examples in Rico~\cite{enrico}. Thus, we create a new high quality dataset by manually relabelling the most common 29 icons on the UI screenshots in Rico. Our correction procedure adds 40\% more icons which reveals the incomplete nature of view hierarchies. We conduct experiments to evaluate the proposed approaches and analyze the results quantitatively and qualitatively. Our proposed model achieves an F1 score of 83\% while the best classification model, which has an architecture similar to ~\cite{Liu:2018}, achieves an F1 score of 55\%. By adding view hierarchy features to the model input, the model's performance increases by 4\% F1 score compared to the pixel only object detection model. 
We hope that our approach can provide insights to other UI related applications and inspire new UI design applications.


\section{Related Work}\label{sec:related_work}
\subsection{Mobile UI Understanding}
With the rapid development of mobile apps and deep learning techniques, there are an increasing number of researches utilizing machine learning to understand UI for a variety of data-driven applications, such as UI semantic analysis, sensitive data detection, and user perception prediction~\cite{supor, uipicker, uiref, wu2019}. We highlight some work about UI element analysis and clarify differences between our work and theirs.

\citet{Liu:2018} proposed an automatic approach for generating
semantic annotations for mobile app UIs. Though they mainly used code based heuristics, they failed to create one for distinguishing icons from the image component and trained a CNN classifier to classify them. The input to the CNN classifier is the icon image cropped from screenshot based on its bounding box denoted in the view hierarchy. 
\citet{Xiao:2019} also adopted a pixel based approach to understand icons. They aimed to identify the sensitive icons that can potentially involve using or collecting users’ sensitive data. For icons that contain texts, they first proposed a set of keywords for each sensitive category and used the edit distance of OCR features detected from the icon with the keywords to classify them. For the other image icons that don't contain text, they utilized a modified version of SIFT~\cite{sift}, a state-of-the-art image feature engineering technique to identify their classes. As an extension of the previous model that classifies icons into sensitive classes for suspicious app detection \cite{Xiao:2019} , \citet{Xi:2019} proposed a joint model for icon classification. They used CNN and LSTM to separately encode image features and text features which are extracted from source code and OCR. Then, a co-attention layer is used to combine them to predict the class. 

Besides the above icon classification models, \citet{chen2020object} is another relevant work to us in that they also frame the UI element annotation task as a domain-specific object detection task. They conducted empirical study of seven representative computer vision-based object detection methods to study their capabilities and limitations.
Compared to them, we propose a novel multimodal object detection model that leverages the resource-id and class name attributes of view hierarchies for understanding icons besides utilizing their visual features.

\subsection{Website UI Understanding}
In parallel with the mobile UI understanding, there are also other works that focus on UIs on the websites. For example, \citet{Kumar2013} built Webzeitgeist, a platform for large-scale design mining comprising a repository of over 100,000 web pages and 100 million design elements. \citet{Lim2012} trained an Support Vector Machine model to classify web page elements into semantic categories such as menu, navigation bar, and title.

\subsection{Object Detection}
As the object detection technique is integral to our work, we briefly review the recent advances in the object detection field. Readers interested in an in-depth review are encouraged to refer to \cite{xiao2020review}. 

The state-of-the-art object detection models in recent years can be categorized into two categories: anchor-based approaches and keypoint-based approaches.
As the representative anchor-based model, SSD~\cite{Liu_2016_ssd} defines a set of anchor boxes with different sizes and uses features extracted from different convolutional feature layers to classify and regress anchor boxes. On the contrary, CenterNet~\cite{zhou2019objects} is a keypoint-based detector that utilizes CNN as a backbone to generate heatmaps, whose peaks correspond to object centers, object bounding box size, and its offsets. They have been widely used in detecting objects on natural images \cite{Lin_2014} and we propose a multi-modal approach based on them to detect artificial icons on mobile screenshots, which is, to the best of our knowledge, the first attempt to adapt multimodal object detection techniques to the UI element annotation task.

\begin{figure*}[t]
  \centering
  \hfill
  \begin{subfigure}[b]{0.5\textwidth}
                \centering
                \includegraphics[width=0.45\linewidth]{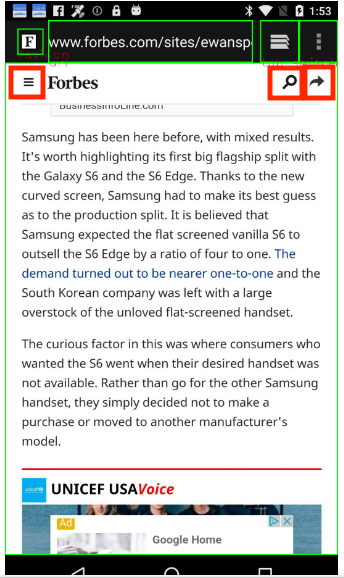}
                \caption{An example when view hierarchy misses nodes.}
                \label{miss_unsync}
        \end{subfigure}%
        \hfill
        \begin{subfigure}[b]{0.5\textwidth}
                \centering
                \includegraphics[width=0.45\linewidth]{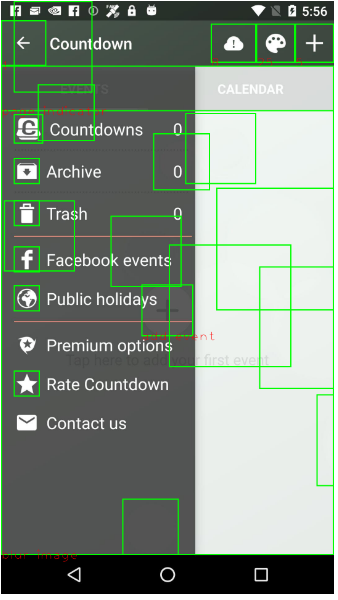}
                \caption{An example when view hierarchy has extra nodes.}
                \label{extra_unsync}
        \end{subfigure}%
        \hfill
  \caption{Examples of unsynchronized screenshot and view hierarchy pairs. The green boxes draws the UI elements denoted in the view hierarchy. The red boxes are the missing UI elements.}
  \label{fig:unsync_example}
\end{figure*}

\section{Dataset}\label{sec:dataset}
We create a high quality UI dataset which contains annotations of the most commonly used icons across 29 classes based on a large-scale mobile design mining dataset - Rico. We refer to the created dataset as Rico$_{clean}$ to distinguish from the original Rico dataset for future reference. In this section, we first illustrate the motivation of our data creation, then the creation process, and lastly, a detailed analysis of the created dataset.

\subsection{Motivation}

The first version of the Rico dataset was collected by \citet{Deka:2017}, it includes 72k screenshots and their view hierarchy. Then \citet{Liu:2018} augmented the data by adding the icon semantic labels to the view hierarchy nodes. To create a taxonomy of icons, they first create a representative icon set by extracting examples based on view hierarchy classes and bounding boxes from Rico. They only kept the examples whose class is ``ImageView" in view hierarchy, whose area is less than 5\% of the total screen area, and whose aspect ratio is greater than 0.75 (smaller dimension divided by larger dimension). Then they hired annotators to annotate the extracted potential icon examples and trained a CNN-based classification model. The ultimate icon labels in Rico are annotated with this model.

There are three problems with this process. First is that the prediction of the CNN model is not necessarily correct. Second is that the criteria they utilized to generate the icon examples for creating the taxonomy inevitably misses some potential icons and adds bias to the machine learning model trained on it. Third is that due to the inconsistency between the render-time of a UI and its view hierarchy, it's possible that the screenshot and its view hierarchy are unsynchronized. It results in inaccurate view hierarchy with missing or excess UI elements. Figure \ref{fig:unsync_example} shows one such example, where view hierarchy missed icon menu, search, and refresh. According to an investigation by \citet{li2020-mapping}, only 25K out of the unique 72K screenshots in Rico have synchronized view hierarchy.

Though suffering from the above issues, Rico dataset is still very valuable as it is the largest mobile app design mining dataset and facilitates many data-driven applications in the HCI field. These motivate us to re-annotate the Rico data and our annotations are based on human perception of the screenshots instead of the view hierarchies.

\subsection{Creation}
We hire 40 in-house professional crowd workers and internal data annotation platform to annotate the most commonly used icons (both bounding boxes and labels) on all the unique 72k UI screenshots in Rico. Crowd workers are given detailed instructions with good and bad annotation examples before they initiate the annotation. They are presented with only the screenshot and required to annotate the bounding box of icons on the screenshot and select their labels from the 29 classes. Our definition of the icon class is based on both the daily app using experience and the definition in the Rico dataset. Note that we add two new common icon classes ``end\_call" and ``take\_photo" in addition to the 27 common icon classes that exist in Rico. On average, crowd workers spend 81 seconds annotating each screenshot.

After the initial annotation, we manually checked the quality of around 300 randomly sampled UIs and summarize the annotation errors. We refine our annotation instruction based on our findings and instruct the same group of the people to verify the annotations and re-annotate the unqualified ones. We repeat this procedure for another two rounds to achieve the final data. 

Through our work, not only are the labels of the existing icons in the original Rico corrected but also new icons that are not captured in the view hierarchy are added. Overall, 51.4\% of the UI elements that are of type icon in the original Rico dataset have their labels corrected and the number of icons after the correction is 40\% more than that in the original Rico. 

\subsection{Analysis}\label{sec:data_analysis}

\begin{figure}[t]
  \centering
  \includegraphics[width=0.9\linewidth]{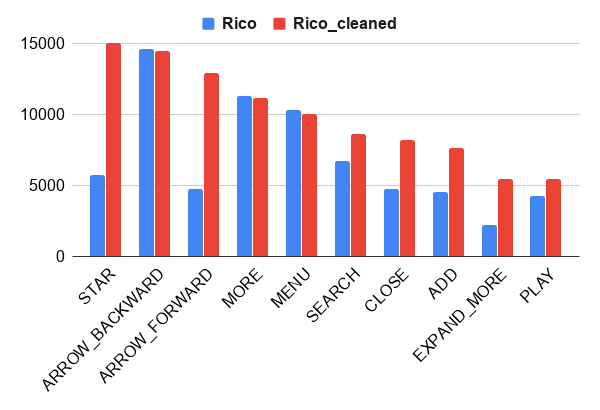}
  \caption{Count of the top 10 frequent icons in Rico and Rico$_{clean}$ datasets. X axis is the icon names. Y axis is the counts of icons. }
  \label{fig:distribution_10}
\end{figure}

\begin{figure*}[t]
  \centering
  \includegraphics[width=0.8\linewidth]{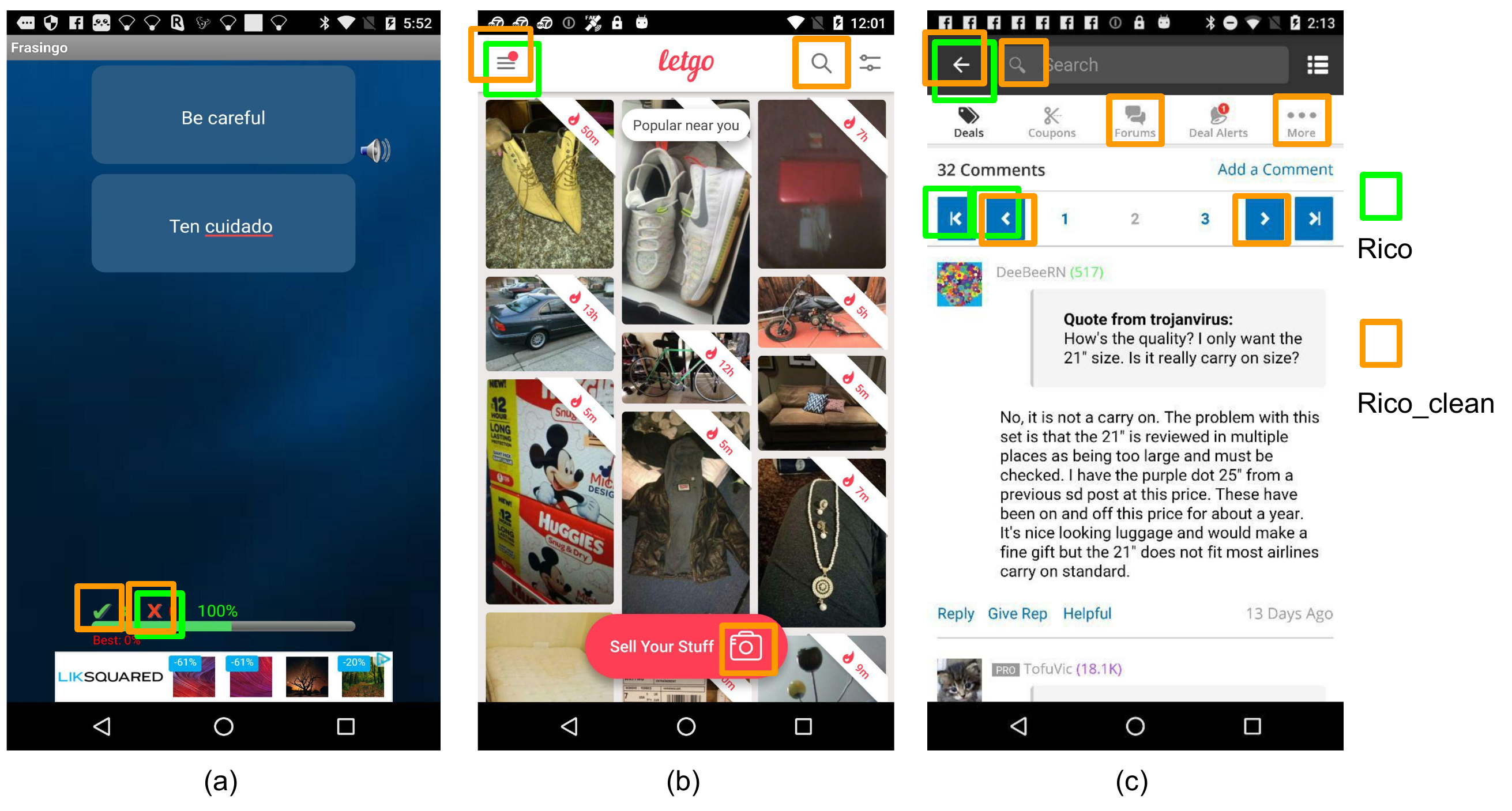}
  \caption{Different annotations in Rico and Rico$_{clean}$. Icons annotated in Rico are marked in green squares, whereas in orange squares for Rico$_{clean}$. Labels of some icons are missing in Rico in all three images. In (c), the icon with shape similar to ``$\mid <$" beneath \textit{arrow backward} icon at the top left is mistakenly labeled as \textit{arrow backward} in Rico and is corrected in Rico$_{clean}$.}
  \label{fig:data_difference}
\end{figure*}

Table~\ref{tab:classes} shows the number of icons and in 29 classes and their examples. Figure \ref{fig:distribution_10} shows the number of the top 10 frequent icons in Rico$_{clean}$ and Rico. We observe that approximately half of the \textit{star}, \textit{arrow forward}, and \textit{expand more} icons missing in Rico are added in Rico$_{clean}$.
Figure~\ref{fig:data_difference} shows the different annotations of the same screenshots in Rico and Rico$_{clean}$ where bounding boxes in Rico is marked in green and bounding boxes in Rico$_{clean}$ is marked in orange. We observe that in Rico, the label of \textit{check}, \textit{search} icons are missing (Figure \ref{fig:data_difference}(a), (b)) and one icon with shape similar to ``$\mid <$" is mistakenly labelled as \textit{arrow backward} (Figure~\ref{fig:data_difference}(c)). Such errors are corrected in Rico$_{clean}$.

\begin{table}
  \caption{Number of icons across 29 classes in Rico$_{clean}$.}
  \label{tab:classes}
  \resizebox{\linewidth}{!}{%
  \begin{tabular}{c|c|c|c|c|c}
   \toprule
Class & Num & Examples & Class & Num & Examples \\ \midrule
star &  15,013 &  \includegraphics[width=0.2in]{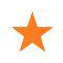}
& home &  2,471 & 
\includegraphics[width=0.2in]{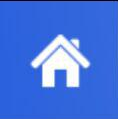}\\\hline
arrow backward  & 14,460 & \includegraphics[width=0.2in]{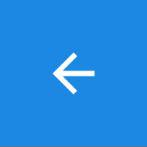}
& refresh & 2,388 & \includegraphics[width=0.2in]{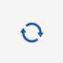}\\\hline
arrow forward & 12,921 & \includegraphics[width=0.2in]{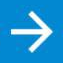}
& time & 2,175 &  \includegraphics[width=0.2in]{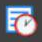}\\\hline
more & 11,169 &  \includegraphics[width=0.2in]{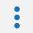}
& emoji & 1,843 &  \includegraphics[width=0.2in]{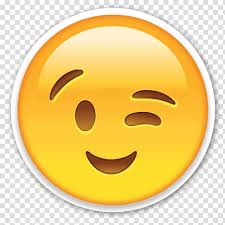}\\\hline
menu &  10,019  & \includegraphics[width=0.2in]{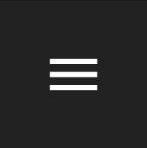}
& edit &  1,528 & \includegraphics[width=0.2in]{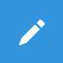} \\\hline
search &  8,618  & \includegraphics[width=0.2in]{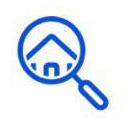}
& notifications &  1,480  & \includegraphics[width=0.2in]{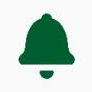}\\\hline
close &  8,196  & \includegraphics[width=0.2in]{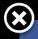}
& call &  1,175  & \includegraphics[width=0.2in]{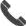} \\\hline
add &  7,654  & \includegraphics[width=0.2in]{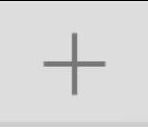}
& pause & 654 & \includegraphics[width=0.2in]{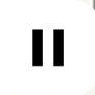} \\\hline
expand more &  5 463  & \includegraphics[width=0.2in]{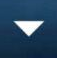}
& send & 643 & \includegraphics[width=0.2in]{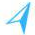}\\\hline
play &  5,444  & \includegraphics[width=0.2in]{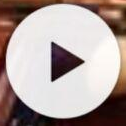}
& delete & 517 & \includegraphics[width=0.2in]{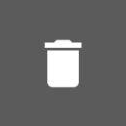} \\\hline
check &  5,189  & \includegraphics[width=0.2in]{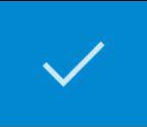}
& video cam & 376 & \includegraphics[width=0.2in]{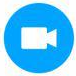} \\\hline
share &  4,115  & \includegraphics[width=0.2in]{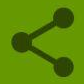}
& launch & 335 & \includegraphics[width=0.2in]{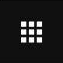}\\\hline
chat &  4,099  &  \includegraphics[width=0.2in]{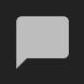}
& end call & 31 & \includegraphics[width=0.2in]{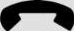} \\\hline
settings &  3,652  &  \includegraphics[width=0.2in]{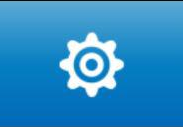}
&take photo & 1,508 & \includegraphics[width=0.2in]{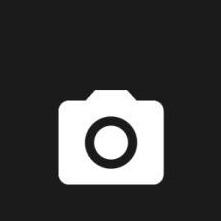} \\\hline
info &  2,618 & \includegraphics[width=0.2in]{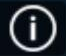}
&  &  &\\
\bottomrule
\end{tabular}%
}

\end{table}

\section{Modeling icon annotation}\label{sec:modeling}
In this section, we first clarify the task definition, then introduce the baseline and the proposing approach in details.
\subsection{Task Definition}\label{sec:task_definition}
 The icon annotation task is defined as: given a user interface represented by a screenshot and view hierarchy pair, one is expected to detect all the icons that belong to a set of pre-defined classes on the UI. 
More specifically, the output contains both icon locations (i.e. coordinates of its bounding boxes) and its class label. 


\subsection{Baseline Classification Model}\label{sec:baseline}

Similar to \citet{Liu:2018}, our baseline classification models first use the locations denoted in view hierarchies as the location of icons and then predict icon types with CNN classification models.

\begin{figure}[t]
  \centering
  \includegraphics[width=0.8\linewidth]{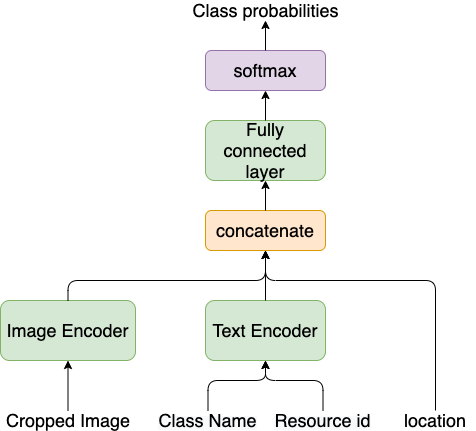}
  \caption{Architecture of the baseline classification model.}
  \label{fig:classify_model}
\end{figure}

In the localization stage, we localize the candidates using the bounding boxes of every leaf nodes in the view hierarchies.
In the classification stage, a machine learning model predicts the label of each candidate from the pre-defined icon set or \textit{OTHER} if the candidate does not belong to any of the pre-defined classes. 
We experiment with two classification models, pixel only and multimodal. The pixel only model is similar to \citet{Liu:2018} which is a CNN-based model taking pixel values as input. Our multimodal classification model is similar to \citet{Xi:2019}, which consists of three deep learning modules: image encoder, text encoder, and feature fusion layer. 
Its complete architecture is summarized in Figure~\ref{fig:classify_model}. 

Our text feature comes from the last text component of class names and resource-ids of view hierarchies. A pre-trained text encoder is used to obtain text embeddings of text features. As this is a shared component with our proposing multimodal object detection model, we elaborate its details in Section \ref{sec:details}.
To obtain the image feature, we first crop the image based on candidate's bounding boxes and rescale it to $225\times225\times3$ in RGB. We also try grayscale images but preliminary experiments doesn't show any improvement in the result. We use the final layer of a pre-trained Inception V2~\cite{Szegedy_2016} to generate image embeddings of the rescaled images, which is followed by a two fully-connected layers with hidden dimensions of 1024 and 128. On top of it, the $relu$ activation function is used to generate the image feature embeddings. 

 To combine the features, we first concatenate the image and text embeddings, along with the location features of the icon, and then run the joint embeddings through a fully connected layer of dimension 128, followed by a softmax function to predict the probability of every class. The choice of all dimension settings are based on the experimental results on the development set.

One problem of the classification model is that as it only predicts labels of the UI elements in a view hierarchy, if the view hierarchy and screenshot are unsynchronized, it experiences loss of recall. On the contrary, the object detection model doesn't suffer from such issue.

\subsection{Detection Model}\label{sec:ours}
\begin{figure*}[t!]
  \centering
  \includegraphics[width=\linewidth]{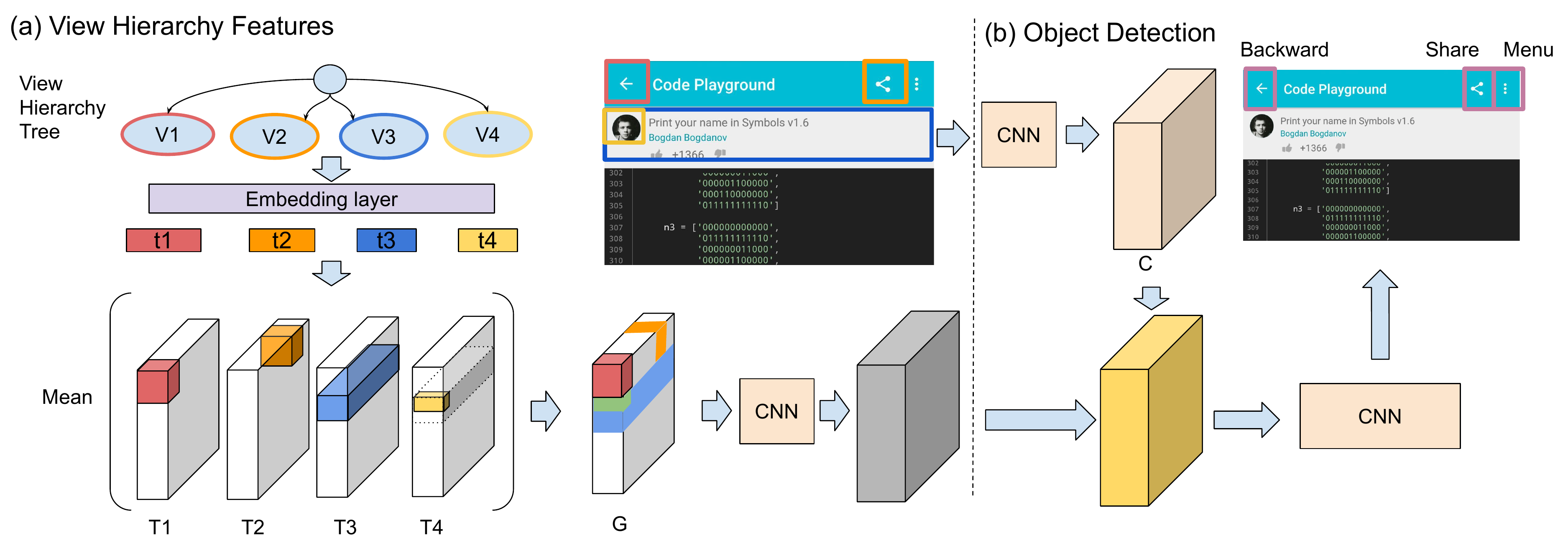}

  \caption{Architecture of the proposed object detection model. (a) shows the computation of view hierarchy feature map. (b) shows how view hierarchy feature is fused to image features.}
  \label{fig:obj_det}
\end{figure*}
Compared with the classification models, object detection models conduct localization and classification in one step. Our proposing object detection model is built upon two state-of-the-art object detection models: SSD~\cite{Liu_2016_ssd} and CenterNet~\cite{zhou2019objects} which are featured with good performance and fast inference speed. Both approaches first generate feature map by a certain CNN. SSD is an anchor-based model that regresses the location and classifies the category of a set of pre-defined anchor boxes to predict objects. On the other hand, CenterNet is anchor free that it predicts the heatmaps whose peaks represent centers, bounding box size, and offsets of the objects for each category from the feature map.

One of our main contributions is to augment the object detection approach to incorporate features from view hierarchy and associate useful information in the view hierarchy leaf nodes to the related image region. Specifically, inspired by the approach in \citep{Wichers_2018}, we propose to build view hierarchy feature map which encodes both view hierarchy nodes' location and textual features. The entire model architecture is illustrated in Figure~\ref{fig:obj_det}.


Given a screenshot, let $\{V_i| i \in [1, S]\}$ be all its view hierarchy leaf nodes, where $S$ is the number of view hierarchy nodes and $V_i$ consists of a textual attribute (i.e. texts in the ``class name" and ``resource-id" fields) and its normalized bounding box coordinate $B_i = (x_{min}, y_{min}, x_{max}, y_{max})$ where $(x_{min}, y_{min})$ and $(x_{max}, y_{max})$ are the top-left and bottom-right coordinates. Let $C\in \mathbb{R}^{H \times W \times D}$ be an intermediate feature map of the object detection model that takes the UI screenshot as the input, where $H=height$, $W=width$, and $D=dimension$. More concretely, in Figure~\ref{fig:obj_det}(b), $C$ is represented by the cube after the square that is marked with ``CNN" and follows the input image. Our next step is to generate a similar feature map for the view hierarchy so that it is easy to do computations to fuse the image and view hierarchy features. 
To achieve it, we use a text encoder to obtain text embedding $t_i\in\mathbb{R}^{K}$ for each view hierarchy leaf node $V_i$ from its textual attributes. Then, using $t_i$, we create a new feature map $T_i$ which is of the same height and width as the image feature map $C$ as follows:
\begin{align}\label{eq:1}
\begin{split}
    O_i = calcOverlay(B_i, H, W, K), \\ 
    T_i = O_i \circ tile(t_i, H, W),
\end{split}
\end{align}
where \textit{calcOverlay} generates $O_i \in \mathbb{R}^{H\times W \times K}$, which is a matrix with only 0 or 1 in each position by computing whether that position is overlayed by the bounding box $B_i$. Specifically, $O_i[p, q, :] = 1$ when $(y_{min} \times H) \leq p < (y_{max} \times H)$ and $(x_{min} \times W) \leq q < (x_{max} \times W)$, and otherwise 0. In equation~\ref{eq:1}, \textit{tile} replicates $t_i$ over height and width dimensions to create a 3-dimensional feature map with the same shape as $O_i$, and $\circ$ denotes the element-wise multiplication.
In other words, $T_i \in \mathbb{R}^{H\times W \times K}$ is filled with the embedding $t_i$ wherever one of the locations overlaps with the bounding box, and is zero elsewhere. This is represented by the cubes at the bottom left of Figure~\ref{fig:obj_det}(a) where the colored part denotes the tiled text embeddings.
  We then compute the average of all $T_i$s to achieve the final feature map $G \in \mathbb{R}^{H\times W \times K}$ by:
\begin{align}\label{eq:2}
\begin{split}
    E = stack([O_1,O_2, ..., O_S]), \\ 
    G = \frac{\sum_{i=0}^{S}T_i}{mean(E)},
\end{split}
\end{align}
where $E \in {R}^{H\times W \times K \times S}$ is generated by stacking all the $O_i$s by its last dimension and \textit{mean} computes the average value along the last dimension of $E$ to generate a matrix of size ${H\times W \times K}$. 
In this way, view hierarchy features are densely associated to its corresponding image region. Figure \ref{fig:obj_det}(a) provides an intuitive visualization of the implementation.

Then, the model converts the dimensionality of each individual vector in $G$ from $K$ to $D$, which is the dimension of the image feature to combine with, by using two $1\times1$ CNN layers. View hierarchy features and image features are combined by addition and this result replaces the original image feature layer in the original object detection model.

\section{Experiments and Results}\label{sec:exp}
This section describes the evaluation, implementation details, quantitative and qualitative analysis of the model performance. 
\subsection{Evaluation Metrics}
We use micro precision, recall, and F1 score as our evaluation metrics, as formulated below. 
\begin{equation}\label{eq:eval}
\begin{aligned}
Precision = \frac{positive\ icon\ prediction}{all\ icon\ prediction}, \\ 
Recall = \frac{positive\ icon\ prediction}{all\ ground\ truth\ icon}, \\
F1 = \frac{2 \cdot precision\cdot recall}{precision+ recall}.
\end{aligned}
\end{equation}

To compute F1 score for object detection models, we first discard all the detection bounding boxes whose confidence score is less than 0.2. Then each detection bounding box is optionally matched to a ground truth icon's bounding box. A detection bounding box is matched if its center is inside the boundaries of the ground truth bounding box. When there are many matches, detection with the highest confidence score is picked. Note that all the matched and unmatched detection bounding boxes are used to compute the F1 score for a fair comparison with the classification model.


\subsection{Implementation Details}\label{sec:details}
Among the variety of properties view hierarchy contains as shown in Figure \ref{fig:screenshot_example}, we utilize the last text component of class names and resource-ids as text features because they provide meaningful hints relevant to the semantics of elements. We tokenize them using underscores and camel-case to obtain the text input to the models. For example, the class name ``\texttt{android.support.AppImageButton}" and resource id ``\texttt{com.sololearn.python:id/vote\_down}" are respectively processed into ``\texttt{app image button}" and ``\texttt{vote down}". 
In all our experiments, we use the pretrained TF Hub nnlm-128 module as the text encoder\footnote{https://tfhub.dev/google/nnlm-en-dim128/1}. The model is a three hidden layer feed-forward Neural-Net Language Models (NN-LM) trained on English Google News 200 billion corpus. The model has around 124 million parameters and maps text to 128-dimensional embedding vectors.

During the training of the baseline, we found resource-id only occurs in 70\% of the icon data in Rico, resulting in empty text feature inputs for many examples. To make full use of text features, we leverage icon labels in the training data as anchor points for aligning image and text feature. Then, we build a dictionary of resource-id text for every icon class. Whenever the resource-id is empty for an icon in the training data, we sample a resource-id from the dictionary of the class based on resource-ids’ frequency. 
Our hypothesis is that randomly sampling a fake resource-id increases the probability of a joint learning between two modalities and enhances the generalizability of the model, which helps to improve the model performance. It is proved by our experimental results in Table~\ref{tab:F1}, which is discussed in the later section. For future reference, this technique is referred to as ``sampling".

We use stacked hourglass \cite{Newell_2016} as the backbone for the CenterNet that takes image of $384\times768$ in RGB and use MobileNet V2 as the feature extractor \cite{Sandler_2018} for SSD that takes image of $448\times730$ in RGB as the majority of images are landscape. 
For CenterNet, we use the final output of the backbone CNN model as the intermediate image feature map, i.e. $C$ in Figure~\ref{fig:obj_det}. For SSD, we experiment with different intermediate feature layers and report the best result. We conduct an ablation study by removing the view hierarchy features in our proposing model architecture to evaluate the affect of using view hierarchy.
\subsection{Quantitative evaluation}
Table~\ref{tab:F1} shows our experimental results in precision, recall and F1 score. To evaluate the contribution of view hierarchy and compare with the prior vision-based approaches, we conduct experiments using only the visual features as input. \textit{Baseline (image only)}, \textit{SSD}, and \textit{CenterNet} use only image inputs, whereas the others use both image and view hierarchy. 

Note that as Rico dataset greatly suffers from the unsynchronization issue between view hierarchy and image that view hierarchy doesn't precisely describe the UI (Section \ref{sec:data_analysis}), to only evaluate the approaches from the modeling perspective and remove the negative effect of this issue on the classification model as it depends on the view hierarchy for element localization, we only use the annotated icons that can be mapped back to a node in the view hierarchy as the ground truth icons when evaluating classification models. In other words, when computing the recall in Equation~\ref{eq:eval}, we only use the icons which are captured in the view hierarchy as the denominator. It is smaller than the total number of annotated icons in Rico$_{clean}$. This setting is under the assumption of a precise view hierarchy. Results computed under this assumption are marked with star in Table \ref{tab:F1}.
We also provide the number when the classification model is evaluated with the same ground truth as the object detection models, which is the fourth row in Table \ref{tab:F1}. We observe that no matter we remove the effect of the unsynchronization issue or not, the best object detection model (\textit{CenterNet+VH}) greatly outperforms the classification models. Especially, \textit{Centernet + VH} beats \textit{baseline (image + VH + sampling)} in F1 score by 28.4\% and in recall by 49.5\%. However, in terms of precision, \textit{baseline (image + VH + sampling)} performs slightly better than the best object detection model \textit{Centernet + VH} by 3.9\%. Depending on the task requirement, precision may outweigh other metrics and the baseline model may be preferred. One can also experiment with the confidence score threshold that decides which predictions to discard for the object detection models to achieve a higher precision, though at the expense of a lower recall.

\begin{table}[t]
  \caption{Precision, Recall, and F1 scores of different models on Rico$_{clean}$. \textit{Baseline()} denotes the classification model and the others are the object detection models. Rows annotated with * consider the setting where only icons whose bounding box is captured in the view hierarchy are included in the ``all ground truth icon" set when computing the recall in Equation~\ref{eq:eval}. This removes the effect of the localization error brought by the view hierarchy from the classification models.}
  \label{tab:F1}
  \begin{tabular}{l|c|c|c}
    \toprule
    Models & Precision & Recall & F1 \\
    \midrule
    \textit{baseline (image only)*} & 0.780 & 0.580 & 0.665 \\ \hline
    \textit{baseline (image + VH)*} & 0.793 & 0.586 &0.674 \\ \hline
    \textit{baseline (image + VH + sampling)*} & 0.804 & 0.604 & 0.690 \\ \hline
    \textit{baseline (image + VH + sampling)} & \textbf{0.804} & 0.416 & 0.548 \\ \hline
    \textit{SSD} & 0.446 & 0.873 & 0.590  \\ \hline
    \textit{SSD + VH} & 0.496 & 0.878 & 0.634  \\ \hline
    \textit{CenterNet} & 0.725 & 0.876 & 0.793 \\ \hline
    \textit{CenterNet + VH} & 0.765 & \textbf{0.911} & \textbf{0.832} \\ \bottomrule
\end{tabular}
\normalsize
\end{table}

 Moreover, we observe that adding view hierarchy information improves the performance of both object detection and classification models on Rico$_{clean}$ across all the metrics, proving the benefits of the multimodal approaches. For classification models, sampling technique is critical to take advantage of the view hierarchy. \textit{CenterNet+VH} performs the best in terms of recall and F1. Compared to \textit{CenterNet}, using view hierarchy by our proposed method helps to improve the precision, recall, F1 by 4\%, 3.5\%, and 3.9\% respectively. CenterNet based model also beat SSD based models across all metrics, which aligns with the result reported by prior works~\cite{zhou2019objects}. To compare the localization ability between different object detection models, we also present the mean average precision (mAP) at IOU threshold 0.1 and 0.5 as in object detection~\cite{map} in Table~\ref{tab:map}. As shown, the models with view hierarchy features consistently achieve better mAP than pure image-based models, indicating the effectiveness of our fusion techniques. \textit{CenterNet} outperforms \textit{SSD} in mAP@0.1IOU but not in mAP@0.5IOU. We observe that though CenterNet models usually predict the center correctly, they tend to predict bounding boxes of smaller size, which we suspect causes the comparatively low mAP@0.5IOU. 
\begin{table}[t]
  \caption{mAP score of different object detection models on Rico$_{clean}$.}
  \label{tab:map}
  \begin{tabular}{l|c|c}
  \toprule
    Models & mAP@0.1IOU & mAP@0.5IOU  \\
    \midrule
    \textit{SSD} & 0.787 & 0.650   \\ \hline
    \textit{SSD + VH} & 0.810 & 0.665  \\ \hline
    \textit{CenterNet} & 0.863 & 0.592 \\ \hline
    \textit{CenterNet + VH} & 0.903 & 0.602  \\ 
    \bottomrule
\end{tabular}
\end{table}


\begin{figure}[t]
  \centering
  \includegraphics[width=\linewidth]{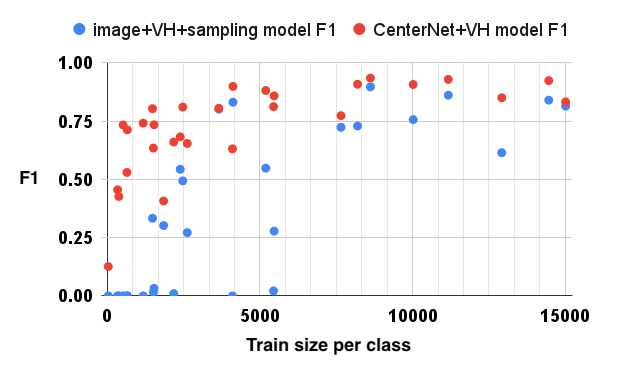}

  \caption{Comparison between the F1 score of the \textit{CenterNet+VH} model and the \textit{baseline (image+VH+sampling)} model for different icon classes on the Rico$_{cleaned}$ dataset. Vertical axis is the F1 score. Horizontal axis is the number of different types of icons in the training set. \textit{CenterNet+VH} in red dots outperforms the baseline model in blue dots especially when the number of the instances in the training set is small.}
  \label{fig:F1Every}
\end{figure}

To understand the results better, for each individual class, we further analyze the relation between the number of instances in the training set with the F1 score evaluated on the test set for the best multimodal classification model (\textit{baseline(image+VH+sampling)}) and the best object detection model (\textit{CenterNet+VH}). Generally, icons with more training data performs better and \textit{CenterNet+VH} performs better across different icon classes. It's worth noticing that our proposed model \textit{CenterNet+VH} outperforms the best classification model especially when a large number of the training data is unavailable.

\subsection{Qualitative evaluation}
\begin{figure}[t]
  \centering
  \includegraphics[width=\linewidth]{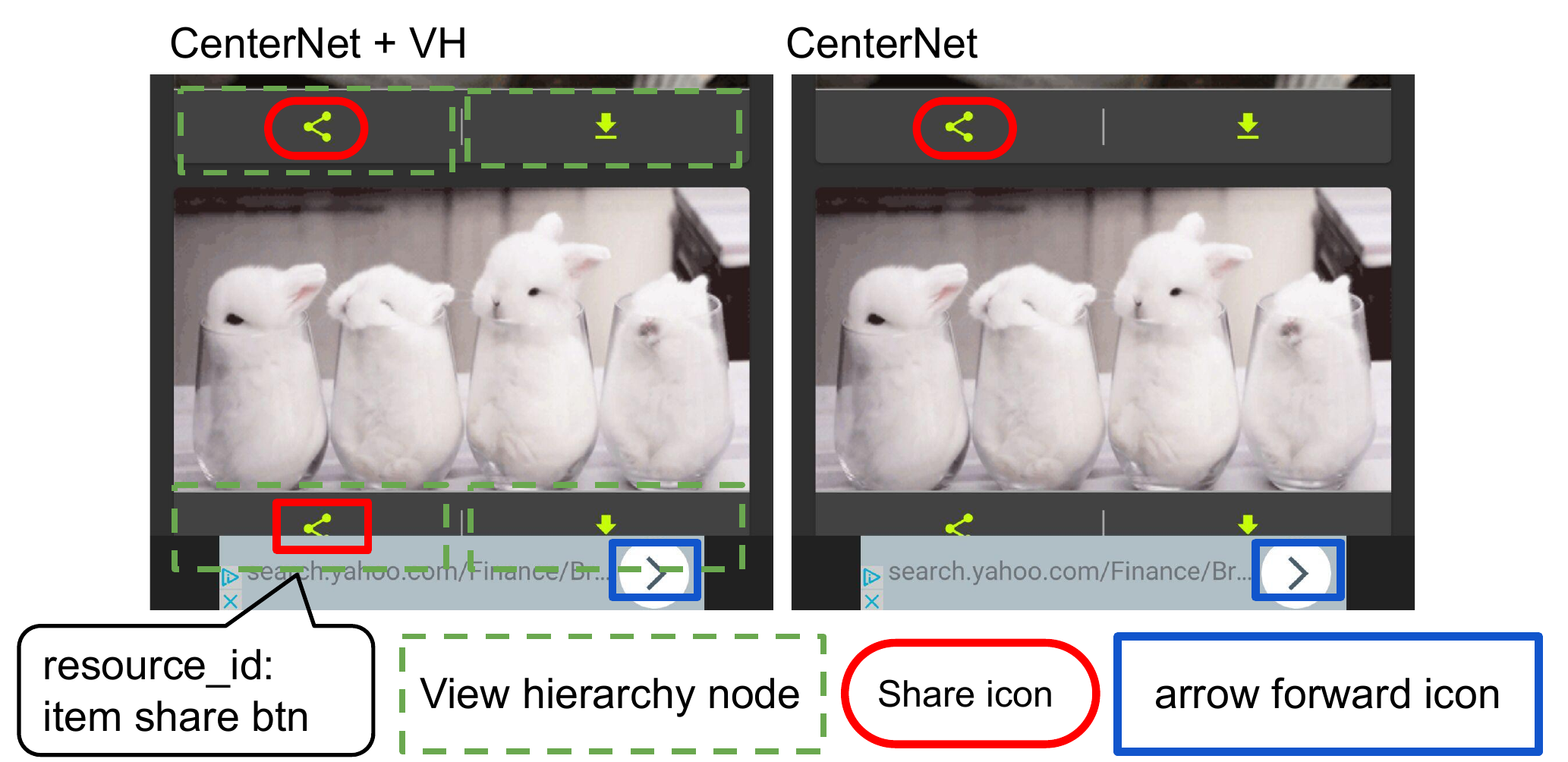}
  \caption{An example showing the effectiveness of using view hierarchy. Prediction results of different classes are marked by squares in different colors and shapes. UI elements that exist in the view hierarchy are drawn in dashed squares.  Left shows the prediction of \textit{CenterNet+VH} and right shows the prediction of \textit{CenterNet}. }
  \label{fig:analysis}
\end{figure}

We also qualitatively compared and analyzed the models performances. Figure~\ref{fig:analysis} shows the predictions of the \textit{CenterNet+VH} and \textit{CenterNet} models on the same screenshot. \textit{Center+VH} correctly predicts the \textit{share icon} located at the bottom of the page even if part of it is occluded by other UI elements. It implies that \textit{CenterNet+VH} model have effectively utilized the view hierarchy feature that contains the ``share" keyword in the resource-id field, whereas a pure image-based model has less clue to make the correct prediction since part of its shape is hidden.

\begin{figure*}[t]
  \centering
  \includegraphics[width=0.9\linewidth]{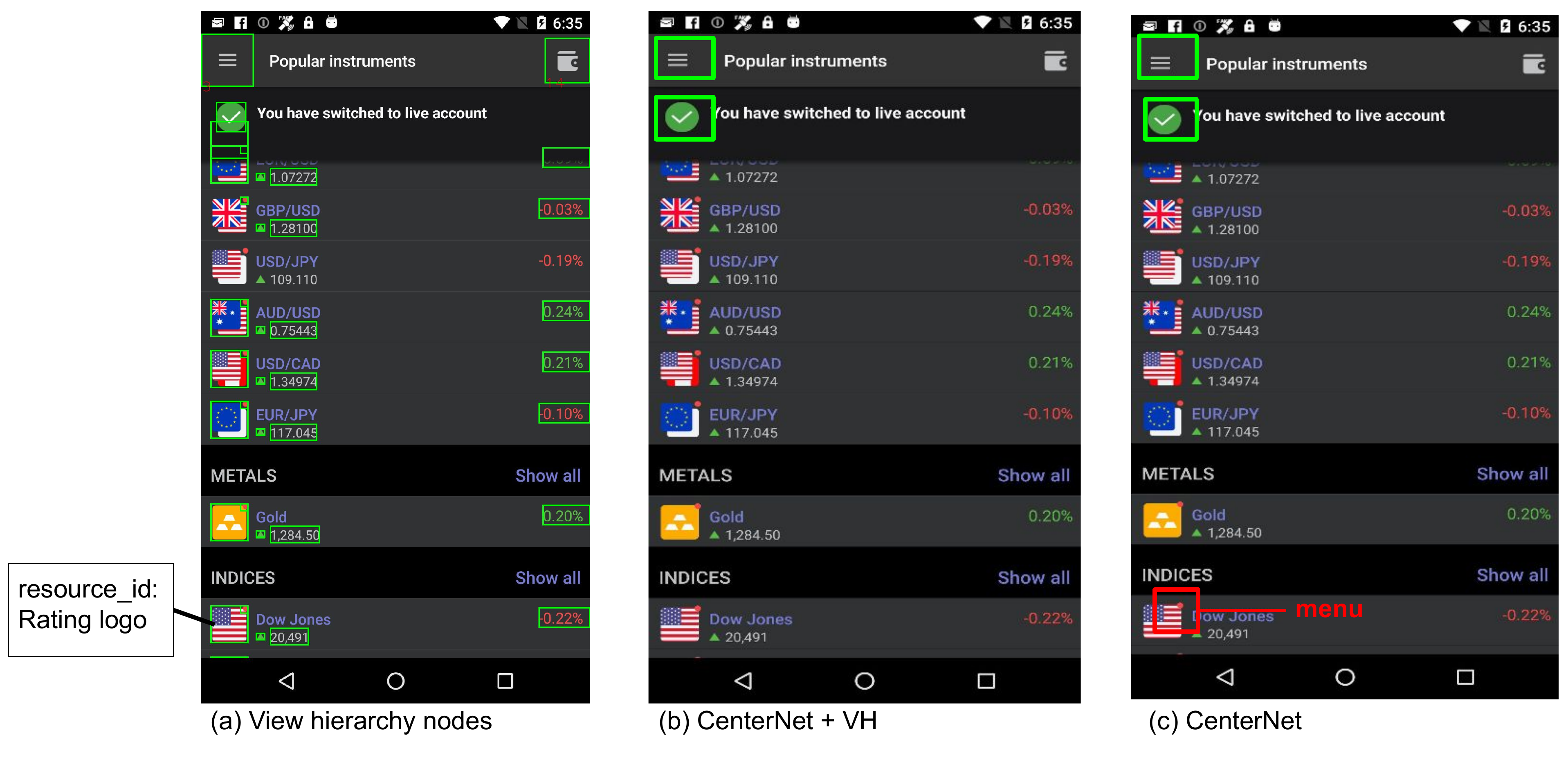}

  \caption{Different predictions by \textit{CenterNet+VH} and \textit{CenterNet} model. (a) is a visualization of view hierarchy leaf nodes. We highlight a part of the view hierarchy features for one node. (b) is the prediction by \textit{CenterNet+VH} model, which matches the ground truth. (c) shows the prediction by \textit{CenterNet}, which mistakenly predicts part of the American flag as \textit{menu} icon.}
  \label{fig:comparison}
\end{figure*}

\begin{figure*}[t]
  \centering
  \includegraphics[width=0.9\linewidth]{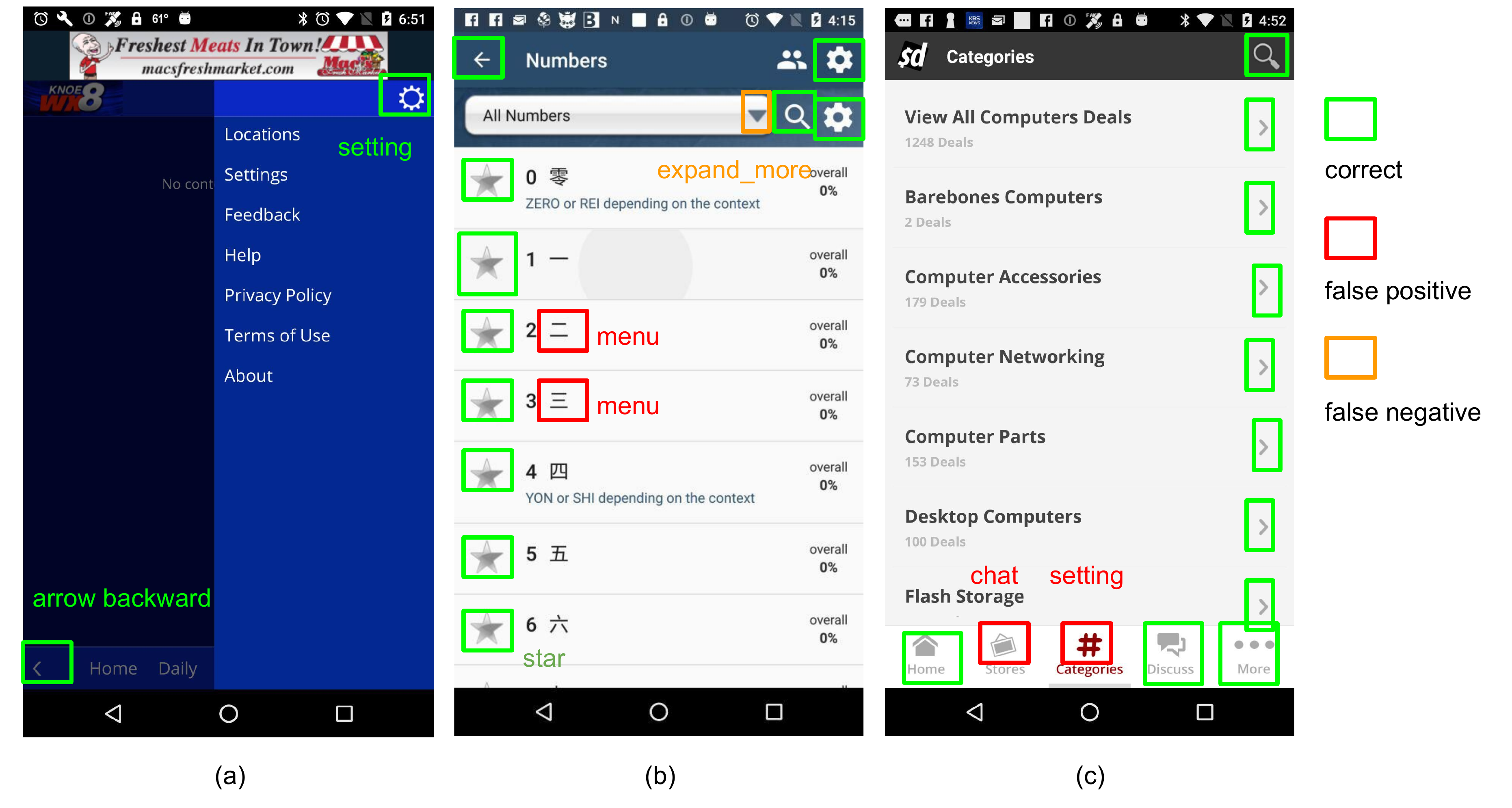}

  \caption{Examples of the predictions by \textit{CenterNet+VH} model. Correct predictions, false positives, and false negatives are separately marked in green, red, and orange. Best viewed in color.}

  \label{fig:difference}
\end{figure*}

Figure~\ref{fig:comparison} shows another example comparing the models using view hierarchy feature and not using view hierarchy feature. A visualization of all the leaf nodes in the view hierarchy is provided in Figure \ref{fig:comparison}(a) with the prediction results of two models in (b) and (c).  The proposed model (\textit{CenterNet+VH}) is able to make correct predictions, whereas the pure image-based model (\textit{CenterNet}) is misled by the shape of the logo and mistakenly predicts the logo with American flag as a \textit{menu} icon. We analyze that the keyword ``logo" in the resource-id of the view hierarchy feature probably have served as a useful hint to the proposed model and help it to make the correct prediction.

Figure~\ref{fig:difference} presents more predictions generated by \textit{CenterNet+VH} model. We observe that though in most cases, our proposed model makes correct predictions, there are also cases where the model failed. In Figure~\ref{fig:difference}(b), the model mistakenly predicts Chinese characters as \textit{menu} icon and in Figure~\ref{fig:difference}(c), ``\#" is predicted as \textit{setting} icon, which indicates that though view hierarchy features help to prevent the model from being biased by the image, it still has trouble distinguishing objects of similar shapes.




\section{Future work}

\paragraph{\textbf{Dataset:}} Currently our dataset only includes commonly used icons. As future work, we will expand the annotations to cover more types of icons. As UI design evolves and the old design becomes outdated, the creation of both a new mobile UI dataset and a new taxonomy of UI elements will be required. However, the proposed deep learning based annotation approach is applicable across all datasets. 

\paragraph{\textbf{Modeling:}} Our current classification baseline represents the state-of-the-art approach, however it only considers the image and text features of the focused UI element while ignoring the contextual information (e.g. entire screenshot and view hierarchy) that the object detection models have access to. A stronger baseline that takes the contextual information as input while maintaining a simple architecture is likely to achieve better performance. One proposal is to first encode the screenshot using a CNN then encode the view hierarchy using a graph neural network before feeding the encoding to the current baseline model. Implementing such a strong baseline to compare with our proposed approach can help demonstrate the benefits presented by the architecture versus just the input features.


\paragraph{\textbf{Application:}} It is easy to expand our proposed icon detection approach to other UI element annotations such as buttons or checkboxes. In addition to UI annotation tasks, we hope this model can provide insights to solve problems in a variety of different domains, such as sensitive icon detection~\cite{Xi:2019}, inferring the user interaction portion of a UI \cite{siva2018}, GUI component detection for automated GUI-prototyping~\cite{k2020}, and tappability prediction \cite{Swearngin_2019}. 
Moreover, our approach is not limited to applications in Android platform and can be easily extended to other platforms (e.g. IOS and website). For example, in the case of websites, our model can take the rendered webpage screenshots along with DOM Trees as input, and output target UI elements. We leave this as another future work.


We expect our trained model to facilitate design based applications for UI designers as well. For example, for every predicted UI element of the trained model, one can obtain its corresponding embeddings from the intermediate layer (output of the backbone network for CenterNet or output of the Region Proposal Networks for SSD) and construct a UI database keyed with such embeddings for UI design search. In such way, given an example UI element on a screenshot, designers can obtain its embeddings and query with it to find similar ones for reference from the UI database. 
In addition to design search, the obtained embeddings can also be used to train a generator model that draws a UI from an arbitrary embedding using Generative Adversarial Network (GAN). The designer can possibly create new UI elements by first doing arithmetic operations on UI embeddings like add or minus and then run the obtained embeddings through a generator model. In this way, designers can easily merge the design essence of two or more UI elements and gain design inspirations.

\section{Conclusion}
UI semantic annotation is beneficial for a number of applications for building an intuitive and pleasant user interaction experience. However, among all types of the UI element, icon labels that represent its functionality are especially hard to achieve since it greatly depends on the visual appearance and can't be automatically generated by heuristics. We created a high-quality dataset with icon annotations by manually re-annotating the most commonly used 29 class of icons in Rico. We built an innovative multimodal icon annotation model based on the state-of-the-art object detection techniques and leveraged both textual features in view hierarchy and image features to solve icon annotation task. Our experimental results on the cleaned dataset show that the proposed model outperforms all the baselines. A qualitative analysis also demonstrates the advantage of our approach over the pure vision-based ones.

Our proposed technology can be directly applied to the accessibility applications by labeling all the icons correctly, which enables the application to read them out and the users to mention them. It also enables querying the same type of icons from a repository of UI data for design mining. As our future work, we hope the proposed approach can also help to solve new tasks like generic UI element annotation, tappability prediction, tasks in other platforms and for design applications. 

\begin{acks}
The authors thank Nevan Wichers, Maria Wang, Gabriel Schubiner, and Lijuan Liu for their guidance and help on dataset creation and processing; Srinivas Sunkara for the assistance with model evaluation; James Stout and Pranav Khaitan for advice, guidance and encouragement; all the anonymous reviewers for reviewing the manuscript and providing valuable feedback.
\end{acks}

\bibliographystyle{ACM-Reference-Format}
\bibliography{sample-base}


\begin{thebibliography}{28}


\ifx \showCODEN    \undefined \def \showCODEN     #1{\unskip}     \fi
\ifx \showDOI      \undefined \def \showDOI       #1{#1}\fi
\ifx \showISBNx    \undefined \def \showISBNx     #1{\unskip}     \fi
\ifx \showISBNxiii \undefined \def \showISBNxiii  #1{\unskip}     \fi
\ifx \showISSN     \undefined \def \showISSN      #1{\unskip}     \fi
\ifx \showLCCN     \undefined \def \showLCCN      #1{\unskip}     \fi
\ifx \shownote     \undefined \def \shownote      #1{#1}          \fi
\ifx \showarticletitle \undefined \def \showarticletitle #1{#1}   \fi
\ifx \showURL      \undefined \def \showURL       {\relax}        \fi
\providecommand\bibfield[2]{#2}
\providecommand\bibinfo[2]{#2}
\providecommand\natexlab[1]{#1}
\providecommand\showeprint[2][]{arXiv:#2}

\bibitem[\protect\citeauthoryear{Andow, Acharya, Li, Enck, Singh, and
  Xie}{Andow et~al\mbox{.}}{2017}]%
        {uiref}
\bibfield{author}{\bibinfo{person}{Benjamin Andow}, \bibinfo{person}{Akhil
  Acharya}, \bibinfo{person}{Dengfeng Li}, \bibinfo{person}{William Enck},
  \bibinfo{person}{Kapil Singh}, {and} \bibinfo{person}{Tao Xie}.}
  \bibinfo{year}{2017}\natexlab{}.
\newblock \showarticletitle{UiRef: Analysis of Sensitive User Inputs in Android
  Applications}. In \bibinfo{booktitle}{\emph{Proceedings of the 10th ACM
  Conference on Security and Privacy in Wireless and Mobile Networks}} (Boston,
  Massachusetts) \emph{(\bibinfo{series}{WiSec ’17})}.
  \bibinfo{publisher}{Association for Computing Machinery},
  \bibinfo{address}{New York, NY, USA}, \bibinfo{pages}{23–34}.
\newblock
\showISBNx{9781450350846}
\urldef\tempurl%
\url{https://doi.org/10.1145/3098243.3098247}
\showDOI{\tempurl}


\bibitem[\protect\citeauthoryear{Chen, Xie, Xing, Chen, Xu, Zhu, and Li}{Chen
  et~al\mbox{.}}{2020}]%
        {chen2020object}
\bibfield{author}{\bibinfo{person}{Jieshan Chen}, \bibinfo{person}{Mulong Xie},
  \bibinfo{person}{Zhenchang Xing}, \bibinfo{person}{Chunyang Chen},
  \bibinfo{person}{Xiwei Xu}, \bibinfo{person}{Liming Zhu}, {and}
  \bibinfo{person}{Guoqiang Li}.} \bibinfo{year}{2020}\natexlab{}.
\newblock \bibinfo{booktitle}{\emph{Object Detection for Graphical User
  Interface: Old Fashioned or Deep Learning or a Combination?}}
\newblock \bibinfo{publisher}{Association for Computing Machinery},
  \bibinfo{address}{New York, NY, USA}, \bibinfo{pages}{1202–1214}.
\newblock
\showISBNx{9781450370431}
\urldef\tempurl%
\url{https://doi.org/10.1145/3368089.3409691}
\showURL{%
\tempurl}


\bibitem[\protect\citeauthoryear{Deka, Huang, Franzen, Hibschman, Afergan, Li,
  Nichols, and Kumar}{Deka et~al\mbox{.}}{2017}]%
        {Deka:2017}
\bibfield{author}{\bibinfo{person}{Biplab Deka}, \bibinfo{person}{Zifeng
  Huang}, \bibinfo{person}{Chad Franzen}, \bibinfo{person}{Joshua Hibschman},
  \bibinfo{person}{Daniel Afergan}, \bibinfo{person}{Yang Li},
  \bibinfo{person}{Jeffrey Nichols}, {and} \bibinfo{person}{Ranjitha Kumar}.}
  \bibinfo{year}{2017}\natexlab{}.
\newblock \showarticletitle{Rico: A Mobile App Dataset for Building Data-Driven
  Design Applications}. In \bibinfo{booktitle}{\emph{Proceedings of the 30th
  Annual Symposium on User Interface Software and Technology}}
  \emph{(\bibinfo{series}{UIST '17})}.
\newblock


\bibitem[\protect\citeauthoryear{Everingham, Eslami, Gool, Williams, Winn, and
  Zisserman}{Everingham et~al\mbox{.}}{2015}]%
        {map}
\bibfield{author}{\bibinfo{person}{Mark Everingham}, \bibinfo{person}{S.~M.
  Eslami}, \bibinfo{person}{Luc Gool}, \bibinfo{person}{Christopher~K.
  Williams}, \bibinfo{person}{John Winn}, {and} \bibinfo{person}{Andrew
  Zisserman}.} \bibinfo{year}{2015}\natexlab{}.
\newblock \showarticletitle{The Pascal Visual Object Classes Challenge: A
  Retrospective}.
\newblock \bibinfo{journal}{\emph{Int. J. Comput. Vision}}
  \bibinfo{volume}{111}, \bibinfo{number}{1} (\bibinfo{date}{Jan.}
  \bibinfo{year}{2015}), \bibinfo{pages}{98–136}.
\newblock
\showISSN{0920-5691}
\urldef\tempurl%
\url{https://doi.org/10.1007/s11263-014-0733-5}
\showDOI{\tempurl}


\bibitem[\protect\citeauthoryear{Huang, Li, Xiao, Wu, Lu, Zhang, and
  Jiang}{Huang et~al\mbox{.}}{2015}]%
        {supor}
\bibfield{author}{\bibinfo{person}{Jianjun Huang}, \bibinfo{person}{Zhichun
  Li}, \bibinfo{person}{Xusheng Xiao}, \bibinfo{person}{Zhenyu Wu},
  \bibinfo{person}{Kangjie Lu}, \bibinfo{person}{Xiangyu Zhang}, {and}
  \bibinfo{person}{Guofei Jiang}.} \bibinfo{year}{2015}\natexlab{}.
\newblock \showarticletitle{{SUPOR}: Precise and Scalable Sensitive User Input
  Detection for Android Apps}. In \bibinfo{booktitle}{\emph{24th {USENIX}
  Security Symposium ({USENIX} Security 15)}}. \bibinfo{publisher}{{USENIX}
  Association}, \bibinfo{address}{Washington, D.C.}, \bibinfo{pages}{977--992}.
\newblock
\showISBNx{978-1-939133-11-3}
\urldef\tempurl%
\url{https://www.usenix.org/conference/usenixsecurity15/technical-sessions/presentation/huang}
\showURL{%
\tempurl}


\bibitem[\protect\citeauthoryear{Kumar, Satyanarayan, Torres, Lim, Ahmad,
  Klemmer, and Talton}{Kumar et~al\mbox{.}}{2013}]%
        {Kumar2013}
\bibfield{author}{\bibinfo{person}{Ranjitha Kumar}, \bibinfo{person}{Arvind
  Satyanarayan}, \bibinfo{person}{Cesar Torres}, \bibinfo{person}{Maxine Lim},
  \bibinfo{person}{Salman Ahmad}, \bibinfo{person}{Scott~R. Klemmer}, {and}
  \bibinfo{person}{Jerry~O. Talton}.} \bibinfo{year}{2013}\natexlab{}.
\newblock \showarticletitle{Webzeitgeist: Design Mining the Web}
  \emph{(\bibinfo{series}{CHI '13})}. \bibinfo{publisher}{Association for
  Computing Machinery}, \bibinfo{address}{New York, NY, USA},
  \bibinfo{pages}{3083–3092}.
\newblock
\showISBNx{9781450318990}
\urldef\tempurl%
\url{https://doi.org/10.1145/2470654.2466420}
\showDOI{\tempurl}


\bibitem[\protect\citeauthoryear{Leiva, Hota, and Oulasvirta}{Leiva
  et~al\mbox{.}}{2020}]%
        {enrico}
\bibfield{author}{\bibinfo{person}{Luis~A. Leiva}, \bibinfo{person}{Asutosh
  Hota}, {and} \bibinfo{person}{Antti Oulasvirta}.}
  \bibinfo{year}{2020}\natexlab{}.
\newblock \showarticletitle{Enrico: A Dataset for Topic Modeling of Mobile UI
  Designs}. In \bibinfo{booktitle}{\emph{22nd International Conference on
  Human-Computer Interaction with Mobile Devices and Services}}
  \emph{(\bibinfo{series}{MobileHCI '20})}. \bibinfo{publisher}{Association for
  Computing Machinery}, Article \bibinfo{articleno}{9},
  \bibinfo{numpages}{4}~pages.
\newblock


\bibitem[\protect\citeauthoryear{Li, He, Zhou, Zhang, and Baldridge}{Li
  et~al\mbox{.}}{2020}]%
        {li2020-mapping}
\bibfield{author}{\bibinfo{person}{Yang Li}, \bibinfo{person}{Jiacong He},
  \bibinfo{person}{Xin Zhou}, \bibinfo{person}{Yuan Zhang}, {and}
  \bibinfo{person}{Jason Baldridge}.} \bibinfo{year}{2020}\natexlab{}.
\newblock \showarticletitle{Mapping Natural Language Instructions to Mobile
  {UI} Action Sequences}. In \bibinfo{booktitle}{\emph{Proceedings of the 58th
  Annual Meeting of the Association for Computational Linguistics}}.
  \bibinfo{publisher}{Association for Computational Linguistics},
  \bibinfo{address}{Online}, \bibinfo{pages}{8198--8210}.
\newblock


\bibitem[\protect\citeauthoryear{Lim, Kumar, Satyanarayan, Torres, Talton, and
  Klemmer}{Lim et~al\mbox{.}}{2 02}]%
        {Lim2012}
\bibfield{author}{\bibinfo{person}{Maxine Lim}, \bibinfo{person}{Ranjitha
  Kumar}, \bibinfo{person}{Arvind Satyanarayan}, \bibinfo{person}{Cesar
  Torres}, \bibinfo{person}{Jerry Talton}, {and} \bibinfo{person}{Scott
  Klemmer}.} \bibinfo{year}{2012-02}\natexlab{}.
\newblock \bibinfo{title}{{Learning Structural Semantics for the Web}}.
\newblock
\newblock
\urldef\tempurl%
\url{http://vis.csail.mit.edu/pubs/web-structural-semantics}
\showURL{%
\tempurl}


\bibitem[\protect\citeauthoryear{Lin, Maire, Belongie, Hays, Perona, Ramanan,
  Dollár, and Zitnick}{Lin et~al\mbox{.}}{2014}]%
        {Lin_2014}
\bibfield{author}{\bibinfo{person}{Tsung-Yi Lin}, \bibinfo{person}{Michael
  Maire}, \bibinfo{person}{Serge Belongie}, \bibinfo{person}{James Hays},
  \bibinfo{person}{Pietro Perona}, \bibinfo{person}{Deva Ramanan},
  \bibinfo{person}{Piotr Dollár}, {and} \bibinfo{person}{C.~Lawrence
  Zitnick}.} \bibinfo{year}{2014}\natexlab{}.
\newblock \showarticletitle{Microsoft COCO: Common Objects in Context}.
\newblock \bibinfo{journal}{\emph{Lecture Notes in Computer Science}}
  (\bibinfo{year}{2014}), \bibinfo{pages}{740–755}.
\newblock
\showISBNx{9783319106021}
\showISSN{1611-3349}
\urldef\tempurl%
\url{https://doi.org/10.1007/978-3-319-10602-1_48}
\showDOI{\tempurl}


\bibitem[\protect\citeauthoryear{Liu, Craft, Situ, Yumer, Mech, and Kumar}{Liu
  et~al\mbox{.}}{2018}]%
        {Liu:2018}
\bibfield{author}{\bibinfo{person}{Thomas~F. Liu}, \bibinfo{person}{Mark
  Craft}, \bibinfo{person}{Jason Situ}, \bibinfo{person}{Ersin Yumer},
  \bibinfo{person}{Radomir Mech}, {and} \bibinfo{person}{Ranjitha Kumar}.}
  \bibinfo{year}{2018}\natexlab{}.
\newblock \showarticletitle{Learning Design Semantics for Mobile Apps}. In
  \bibinfo{booktitle}{\emph{The 31st Annual ACM Symposium on User Interface
  Software and Technology}} (Berlin, Germany) \emph{(\bibinfo{series}{UIST
  '18})}. \bibinfo{publisher}{ACM}, \bibinfo{address}{New York, NY, USA},
  \bibinfo{pages}{569--579}.
\newblock
\showISBNx{978-1-4503-5948-1}
\urldef\tempurl%
\url{https://doi.org/10.1145/3242587.3242650}
\showDOI{\tempurl}


\bibitem[\protect\citeauthoryear{Liu, Anguelov, Erhan, Szegedy, Reed, Fu, and
  Berg}{Liu et~al\mbox{.}}{2016}]%
        {Liu_2016_ssd}
\bibfield{author}{\bibinfo{person}{Wei Liu}, \bibinfo{person}{Dragomir
  Anguelov}, \bibinfo{person}{Dumitru Erhan}, \bibinfo{person}{Christian
  Szegedy}, \bibinfo{person}{Scott Reed}, \bibinfo{person}{Cheng-Yang Fu},
  {and} \bibinfo{person}{Alexander~C. Berg}.} \bibinfo{year}{2016}\natexlab{}.
\newblock \showarticletitle{SSD: Single Shot MultiBox Detector}.
\newblock \bibinfo{journal}{\emph{Lecture Notes in Computer Science}}
  (\bibinfo{year}{2016}), \bibinfo{pages}{21–37}.
\newblock
\showISBNx{9783319464480}
\showISSN{1611-3349}
\urldef\tempurl%
\url{https://doi.org/10.1007/978-3-319-46448-0_2}
\showDOI{\tempurl}


\bibitem[\protect\citeauthoryear{Lowe}{Lowe}{2004}]%
        {sift}
\bibfield{author}{\bibinfo{person}{David Lowe}.}
  \bibinfo{year}{2004}\natexlab{}.
\newblock \showarticletitle{Distinctive Image Features from Scale-Invariant
  Keypoints}.
\newblock \bibinfo{journal}{\emph{International Journal of Computer Vision}}
  \bibinfo{volume}{60} (\bibinfo{date}{11} \bibinfo{year}{2004}),
  \bibinfo{pages}{91--110}.
\newblock
\urldef\tempurl%
\url{https://doi.org/10.1023/B%3AVISI.0000029664.99615.94}
\showDOI{\tempurl}


\bibitem[\protect\citeauthoryear{{Moran}, {Bernal-Cárdenas}, {Curcio},
  {Bonett}, and {Poshyvanyk}}{{Moran} et~al\mbox{.}}{2020}]%
        {k2020}
\bibfield{author}{\bibinfo{person}{K. {Moran}}, \bibinfo{person}{C.
  {Bernal-Cárdenas}}, \bibinfo{person}{M. {Curcio}}, \bibinfo{person}{R.
  {Bonett}}, {and} \bibinfo{person}{D. {Poshyvanyk}}.}
  \bibinfo{year}{2020}\natexlab{}.
\newblock \showarticletitle{Machine Learning-Based Prototyping of Graphical
  User Interfaces for Mobile Apps}.
\newblock \bibinfo{journal}{\emph{IEEE Transactions on Software Engineering}}
  \bibinfo{volume}{46}, \bibinfo{number}{2} (\bibinfo{year}{2020}),
  \bibinfo{pages}{196--221}.
\newblock


\bibitem[\protect\citeauthoryear{Nan, Yang, Yang, Zhou, Gu, and Wang}{Nan
  et~al\mbox{.}}{2015}]%
        {uipicker}
\bibfield{author}{\bibinfo{person}{Yuhong Nan}, \bibinfo{person}{Min Yang},
  \bibinfo{person}{Zhemin Yang}, \bibinfo{person}{Shunfan Zhou},
  \bibinfo{person}{Guofei Gu}, {and} \bibinfo{person}{XiaoFeng Wang}.}
  \bibinfo{year}{2015}\natexlab{}.
\newblock \showarticletitle{UIPicker: User-Input Privacy Identification in
  Mobile Applications}. In \bibinfo{booktitle}{\emph{24th {USENIX} Security
  Symposium ({USENIX} Security 15)}}. \bibinfo{publisher}{{USENIX}
  Association}, \bibinfo{address}{Washington, D.C.},
  \bibinfo{pages}{993--1008}.
\newblock
\showISBNx{978-1-939133-11-3}
\urldef\tempurl%
\url{https://www.usenix.org/conference/usenixsecurity15/technical-sessions/presentation/nan}
\showURL{%
\tempurl}


\bibitem[\protect\citeauthoryear{{Natarajan} and {Csallner}}{{Natarajan} and
  {Csallner}}{2018}]%
        {siva2018}
\bibfield{author}{\bibinfo{person}{S. {Natarajan}} {and} \bibinfo{person}{C.
  {Csallner}}.} \bibinfo{year}{2018}\natexlab{}.
\newblock \showarticletitle{P2A: A Tool for Converting Pixels to Animated
  Mobile Application User Interfaces}. In \bibinfo{booktitle}{\emph{2018
  IEEE/ACM 5th International Conference on Mobile Software Engineering and
  Systems (MOBILESoft)}}. \bibinfo{pages}{224--235}.
\newblock


\bibitem[\protect\citeauthoryear{Newell, Yang, and Deng}{Newell
  et~al\mbox{.}}{2016}]%
        {Newell_2016}
\bibfield{author}{\bibinfo{person}{Alejandro Newell}, \bibinfo{person}{Kaiyu
  Yang}, {and} \bibinfo{person}{Jia Deng}.} \bibinfo{year}{2016}\natexlab{}.
\newblock \showarticletitle{Stacked Hourglass Networks for Human Pose
  Estimation}.
\newblock \bibinfo{journal}{\emph{Lecture Notes in Computer Science}}
  (\bibinfo{year}{2016}), \bibinfo{pages}{483–499}.
\newblock
\showISBNx{9783319464848}
\showISSN{1611-3349}
\urldef\tempurl%
\url{https://doi.org/10.1007/978-3-319-46484-8_29}
\showDOI{\tempurl}


\bibitem[\protect\citeauthoryear{Sandler, Howard, Zhu, Zhmoginov, and
  Chen}{Sandler et~al\mbox{.}}{2018}]%
        {Sandler_2018}
\bibfield{author}{\bibinfo{person}{Mark Sandler}, \bibinfo{person}{Andrew
  Howard}, \bibinfo{person}{Menglong Zhu}, \bibinfo{person}{Andrey Zhmoginov},
  {and} \bibinfo{person}{Liang-Chieh Chen}.} \bibinfo{year}{2018}\natexlab{}.
\newblock \showarticletitle{MobileNetV2: Inverted Residuals and Linear
  Bottlenecks}.
\newblock \bibinfo{journal}{\emph{2018 IEEE/CVF Conference on Computer Vision
  and Pattern Recognition}} (\bibinfo{date}{Jun} \bibinfo{year}{2018}).
\newblock
\showISBNx{9781538664209}
\urldef\tempurl%
\url{https://doi.org/10.1109/cvpr.2018.00474}
\showDOI{\tempurl}


\bibitem[\protect\citeauthoryear{Swearngin and Li}{Swearngin and Li}{2019}]%
        {Swearngin_2019}
\bibfield{author}{\bibinfo{person}{Amanda Swearngin} {and}
  \bibinfo{person}{Yang Li}.} \bibinfo{year}{2019}\natexlab{}.
\newblock \showarticletitle{Modeling Mobile Interface Tappability Using
  Crowdsourcing and Deep Learning}.
\newblock \bibinfo{journal}{\emph{Proceedings of the 2019 CHI Conference on
  Human Factors in Computing Systems - CHI ’19}} (\bibinfo{year}{2019}).
\newblock


\bibitem[\protect\citeauthoryear{Szegedy, Vanhoucke, Ioffe, Shlens, and
  Wojna}{Szegedy et~al\mbox{.}}{2016}]%
        {Szegedy_2016}
\bibfield{author}{\bibinfo{person}{Christian Szegedy}, \bibinfo{person}{Vincent
  Vanhoucke}, \bibinfo{person}{Sergey Ioffe}, \bibinfo{person}{Jon Shlens},
  {and} \bibinfo{person}{Zbigniew Wojna}.} \bibinfo{year}{2016}\natexlab{}.
\newblock \showarticletitle{Rethinking the Inception Architecture for Computer
  Vision}.
\newblock \bibinfo{journal}{\emph{2016 IEEE Conference on Computer Vision and
  Pattern Recognition (CVPR)}} (\bibinfo{date}{Jun} \bibinfo{year}{2016}).
\newblock
\showISBNx{9781467388511}
\urldef\tempurl%
\url{https://doi.org/10.1109/cvpr.2016.308}
\showDOI{\tempurl}


\bibitem[\protect\citeauthoryear{Szpiro, Hashash, Zhao, and Azenkot}{Szpiro
  et~al\mbox{.}}{2016}]%
        {szpiro2016people}
\bibfield{author}{\bibinfo{person}{Sarit Felicia~Anais Szpiro},
  \bibinfo{person}{Shafeka Hashash}, \bibinfo{person}{Yuhang Zhao}, {and}
  \bibinfo{person}{Shiri Azenkot}.} \bibinfo{year}{2016}\natexlab{}.
\newblock \showarticletitle{How people with low vision access computing
  devices: Understanding challenges and opportunities}. In
  \bibinfo{booktitle}{\emph{Proceedings of the 18th International ACM SIGACCESS
  Conference on Computers and Accessibility}}. \bibinfo{pages}{171--180}.
\newblock


\bibitem[\protect\citeauthoryear{Wichers, Hakkani-Tur, and Chen}{Wichers
  et~al\mbox{.}}{2018}]%
        {Wichers_2018}
\bibfield{author}{\bibinfo{person}{Nevan Wichers}, \bibinfo{person}{Dilek
  Hakkani-Tur}, {and} \bibinfo{person}{Jindong Chen}.}
  \bibinfo{year}{2018}\natexlab{}.
\newblock \showarticletitle{Resolving Referring Expressions in Images with
  Labeled Elements}.
\newblock \bibinfo{journal}{\emph{2018 IEEE Spoken Language Technology Workshop
  (SLT)}} (\bibinfo{date}{Dec} \bibinfo{year}{2018}).
\newblock
\showISBNx{9781538643341}
\urldef\tempurl%
\url{https://doi.org/10.1109/slt.2018.8639518}
\showDOI{\tempurl}


\bibitem[\protect\citeauthoryear{Wu, Kim, Li, and Ma}{Wu et~al\mbox{.}}{2019}]%
        {wu2019}
\bibfield{author}{\bibinfo{person}{Ziming Wu}, \bibinfo{person}{Taewook Kim},
  \bibinfo{person}{Quan Li}, {and} \bibinfo{person}{Xiaojuan Ma}.}
  \bibinfo{year}{2019}\natexlab{}.
\newblock \showarticletitle{Understanding and Modeling User-Perceived Brand
  Personality from Mobile Application UIs}. In
  \bibinfo{booktitle}{\emph{Proceedings of the 2019 CHI Conference on Human
  Factors in Computing Systems}} \emph{(\bibinfo{series}{CHI '19})}.
  \bibinfo{pages}{1–12}.
\newblock


\bibitem[\protect\citeauthoryear{Xi, Yang, Xiao, Yao, Xiong, Xu, Wang, Gao,
  Liu, Xu, et~al\mbox{.}}{Xi et~al\mbox{.}}{2019}]%
        {Xi:2019}
\bibfield{author}{\bibinfo{person}{Shengqu Xi}, \bibinfo{person}{Shao Yang},
  \bibinfo{person}{Xusheng Xiao}, \bibinfo{person}{Yuan Yao},
  \bibinfo{person}{Yayuan Xiong}, \bibinfo{person}{Fengyuan Xu},
  \bibinfo{person}{Haoyu Wang}, \bibinfo{person}{Peng Gao},
  \bibinfo{person}{Zhuotao Liu}, \bibinfo{person}{Feng Xu}, {et~al\mbox{.}}}
  \bibinfo{year}{2019}\natexlab{}.
\newblock \showarticletitle{DeepIntent: Deep icon-behavior learning for
  detecting intention-behavior discrepancy in mobile apps}. In
  \bibinfo{booktitle}{\emph{Proceedings of the 2019 ACM SIGSAC Conference on
  Computer and Communications Security}}. \bibinfo{pages}{2421--2436}.
\newblock


\bibitem[\protect\citeauthoryear{Xiao, Wang, Cao, Wang, and Gao}{Xiao
  et~al\mbox{.}}{2019}]%
        {Xiao:2019}
\bibfield{author}{\bibinfo{person}{Xusheng Xiao}, \bibinfo{person}{Xiaoyin
  Wang}, \bibinfo{person}{Zhihao Cao}, \bibinfo{person}{Hanlin Wang}, {and}
  \bibinfo{person}{Peng Gao}.} \bibinfo{year}{2019}\natexlab{}.
\newblock \showarticletitle{IconIntent: Automatic Identification of Sensitive
  UI Widgets Based on Icon Classification for Android Apps}.
  \bibinfo{pages}{257--268}.
\newblock
\urldef\tempurl%
\url{https://doi.org/10.1109/ICSE.2019.00041}
\showDOI{\tempurl}


\bibitem[\protect\citeauthoryear{Xiao, Tian, Yu, Zhang, Liu, Du, and Lan}{Xiao
  et~al\mbox{.}}{2020}]%
        {xiao2020review}
\bibfield{author}{\bibinfo{person}{Youzi Xiao}, \bibinfo{person}{Zhiqiang
  Tian}, \bibinfo{person}{Jiachen Yu}, \bibinfo{person}{Yinshu Zhang},
  \bibinfo{person}{Shuai Liu}, \bibinfo{person}{Shaoyi Du}, {and}
  \bibinfo{person}{Xuguang Lan}.} \bibinfo{year}{2020}\natexlab{}.
\newblock \showarticletitle{A review of object detection based on deep
  learning}.
\newblock \bibinfo{journal}{\emph{Multimedia Tools and Applications}}
  \bibinfo{volume}{79}, \bibinfo{number}{33} (\bibinfo{year}{2020}),
  \bibinfo{pages}{23729--23791}.
\newblock


\bibitem[\protect\citeauthoryear{Zhang, de~Greef, Swearngin, White, Murray, Yu,
  Shan, Nichols, Wu, Fleizach, Everitt, and Bigham}{Zhang
  et~al\mbox{.}}{2021}]%
        {zhang2021screen}
\bibfield{author}{\bibinfo{person}{Xiaoyi Zhang}, \bibinfo{person}{Lilian de
  Greef}, \bibinfo{person}{Amanda Swearngin}, \bibinfo{person}{Samuel White},
  \bibinfo{person}{Kyle Murray}, \bibinfo{person}{Lisa Yu}, \bibinfo{person}{Qi
  Shan}, \bibinfo{person}{Jeffrey Nichols}, \bibinfo{person}{Jason Wu},
  \bibinfo{person}{Chris Fleizach}, \bibinfo{person}{Aaron Everitt}, {and}
  \bibinfo{person}{Jeffrey~P. Bigham}.} \bibinfo{year}{2021}\natexlab{}.
\newblock \bibinfo{title}{Screen Recognition: Creating Accessibility Metadata
  for Mobile Applications from Pixels}.
\newblock
\newblock
\showeprint[arxiv]{2101.04893}~[cs.HC]


\bibitem[\protect\citeauthoryear{Zhou, Wang, and Krähenbühl}{Zhou
  et~al\mbox{.}}{2019}]%
        {zhou2019objects}
\bibfield{author}{\bibinfo{person}{Xingyi Zhou}, \bibinfo{person}{Dequan Wang},
  {and} \bibinfo{person}{Philipp Krähenbühl}.}
  \bibinfo{year}{2019}\natexlab{}.
\newblock \bibinfo{title}{Objects as Points}.
\newblock
\newblock
\showeprint[arxiv]{1904.07850}~[cs.CV]


\end{thebibliography}


\end{document}